\documentclass[11pt]{article}
\usepackage{acl2016}
\usepackage{times}
\usepackage{latexsym}
\usepackage{url}
\usepackage{booktabs}
\usepackage{graphicx}
\usepackage{color}
\usepackage{amsmath}

\aclfinalcopy 
\def\aclpaperid{384} 

\title{Multilingual Part-of-Speech Tagging with\\Bidirectional Long Short-Term Memory Models and Auxiliary Loss}
\author{Barbara Plank \\ University of Groningen \\ The Netherlands\\ {\tt b.plank@rug.nl}\\
         \And  Anders S\o{}gaard\\University of Copenhagen\\Denmark \\ {\tt soegaard@hum.ku.dk} \hspace*{0.2cm}
         \And Yoav Goldberg\\Bar-Ilan University\\Israel\\ {\tt yoav.goldberg@gmail.com}
         }

\date{}

\begin{document}

\maketitle

\begin{abstract}
Bidirectional long short-term memory (bi-LSTM) networks have recently proven successful for various NLP sequence modeling tasks, but little is known about their reliance to input representations, target languages, data set size, and label noise. 
We address these issues and evaluate bi-LSTMs with word, character, and unicode byte embeddings for POS tagging. We compare bi-LSTMs to traditional POS taggers across languages and data sizes. We also present a novel bi-LSTM model, which combines the POS tagging loss function with an auxiliary loss function that accounts for rare words. The model obtains state-of-the-art performance across 22 languages, and works especially well for morphologically complex languages. Our analysis suggests that bi-LSTMs are less sensitive to training data size and label corruptions (at small noise levels) than previously assumed. 
\end{abstract}

\section{Introduction}

Recently, bidirectional long short-term memory networks (bi-LSTM)~\cite{graves:schmidhuber:2005,Hochreiter:Schmidhuber:97} have been used for 
language modelling \cite{ling:ea:2015}, POS tagging \cite{ling:ea:2015,wang:ea:2015:arxiv}, transition-based dependency parsing \cite{ballesteros:ea:2015,kiperwasser:goldberg:2016}, fine-grained sentiment analysis \cite{Liu:ea:15}, syntactic chunking \cite{Huang:ea:15}, and semantic role labeling \cite{Zhou:Xu:15}. LSTMs are recurrent neural networks (RNNs) in which layers are designed to prevent vanishing gradients. Bidirectional LSTMs make a backward and forward pass through the sequence before passing on to the next layer. For further details, see \cite{goldberg-primer,cho-primer}.


We consider using bi-LSTMs for POS tagging. Previous work on using deep learning-based methods for POS tagging has focused either on a single language~\cite{Collobert:ea:2011natural,wang:ea:2015:arxiv} or a small set of languages~\cite{ling:ea:2015,santos:zadrozny:2014}. Instead we evaluate our models across 22 languages.
In addition, we compare performance with representations at different levels of granularity (words, characters, and bytes). 
These levels of representation were previously introduced 
in different efforts \cite{chrupala:2013,zhang:ea:2015:char,ling:ea:2015,santos:zadrozny:2014,gillick:ea:2016,kim2015character}, but a comparative evaluation was missing.

Moreover, deep networks are often said to require large volumes of training data. We investigate to what extent bi-LSTMs are more sensitive to the amount of training data and label noise than 
standard POS taggers.

Finally, we introduce a novel model, a bi-LSTM trained with auxiliary loss. The model jointly predicts the POS and the log frequency of 
the word. The intuition behind this model is that the auxiliary loss, being predictive of word frequency, helps to differentiate the representations of rare and common words.
We indeed observe performance gains on rare and out-of-vocabulary words. These performance gains transfer into general improvements for morphologically rich languages. 

\paragraph{Contributions} In this paper, we a) evaluate the effectiveness of different representations in bi-LSTMs, 
b) compare these models across a large set of languages and under varying conditions (data size, label noise) and
c) propose a novel bi-LSTM model with auxiliary loss (\textsc{Logfreq}). 

\section{Tagging with bi-LSTMs}

Recurrent neural networks (RNNs)~\cite{elman:1990} allow the computation of fixed-size vector representations for word sequences of arbitrary length.  
An RNN is a function that reads in $n$ vectors $x_1,...,x_n$ and produces an output vector $h_n$, that depends on the entire sequence $x_1,...,x_n$. 
The vector $h_n$ is then fed as an input to some classifier, or higher-level RNNs in stacked/hierarchical models.
The entire network is trained jointly such that the hidden representation captures the important information from the sequence for the prediction task.

A bidirectional recurrent neural network (bi-RNN)~\cite{graves:schmidhuber:2005} is an extension of an RNN that reads the input sequence twice, 
from left to right and right to left, and the encodings are concatenated. 
The literature uses the term bi-RNN to refer to two related architectures, which we refer to here as ``context bi-RNN'' and ``sequence bi-RNN''.
In a sequence bi-RNN (bi-RNN$_{\text{seq}}$), the input is a sequence of vectors $x_{1:n}$ and the output is a concatenation ($\circ$) of a forward ($f$) and reverse ($r$) 
RNN each reading the sequence in a different directions:
\[ \scalebox{.95}{$ v = \text{bi-RNN}_\text{seq}(x_{1:n}) = \text{RNN}_f(x_{1:n}) \circ \text{RNN}_r(x_{n:1}) $} \]
In a context bi-RNN (bi-RNN$_{\text{ctx}}$), we get an additional input $i$ indicating a sequence position, and the resulting vectors $v_i$ result from concatenating the RNN encodings up to $i$: 
\[ \scalebox{.93}{$ v_i = \text{bi-RNN}_\text{ctx}(x_{1:n},i) = \text{RNN}_f(x_{1:i}) \circ \text{RNN}_r(x_{n:i}) $} \]
Thus, the state vector $v_i$ in this bi-RNN encodes information 
at position $i$ and its entire sequential context.
Another view of the context bi-RNN is of taking a sequence $x_{1:n}$ and returning the corresponding sequence of state vectors $v_{1:n}$.

LSTMs \cite{Hochreiter:Schmidhuber:97} are a variant of RNNs that replace the cells of RNNs
with LSTM cells that were designed to prevent vanishing gradients. 
Bidirectional LSTMs are the bi-RNN counterpart based on LSTMs.

\begin{figure}[ht!]\centering
\includegraphics[width=1.0\columnwidth]{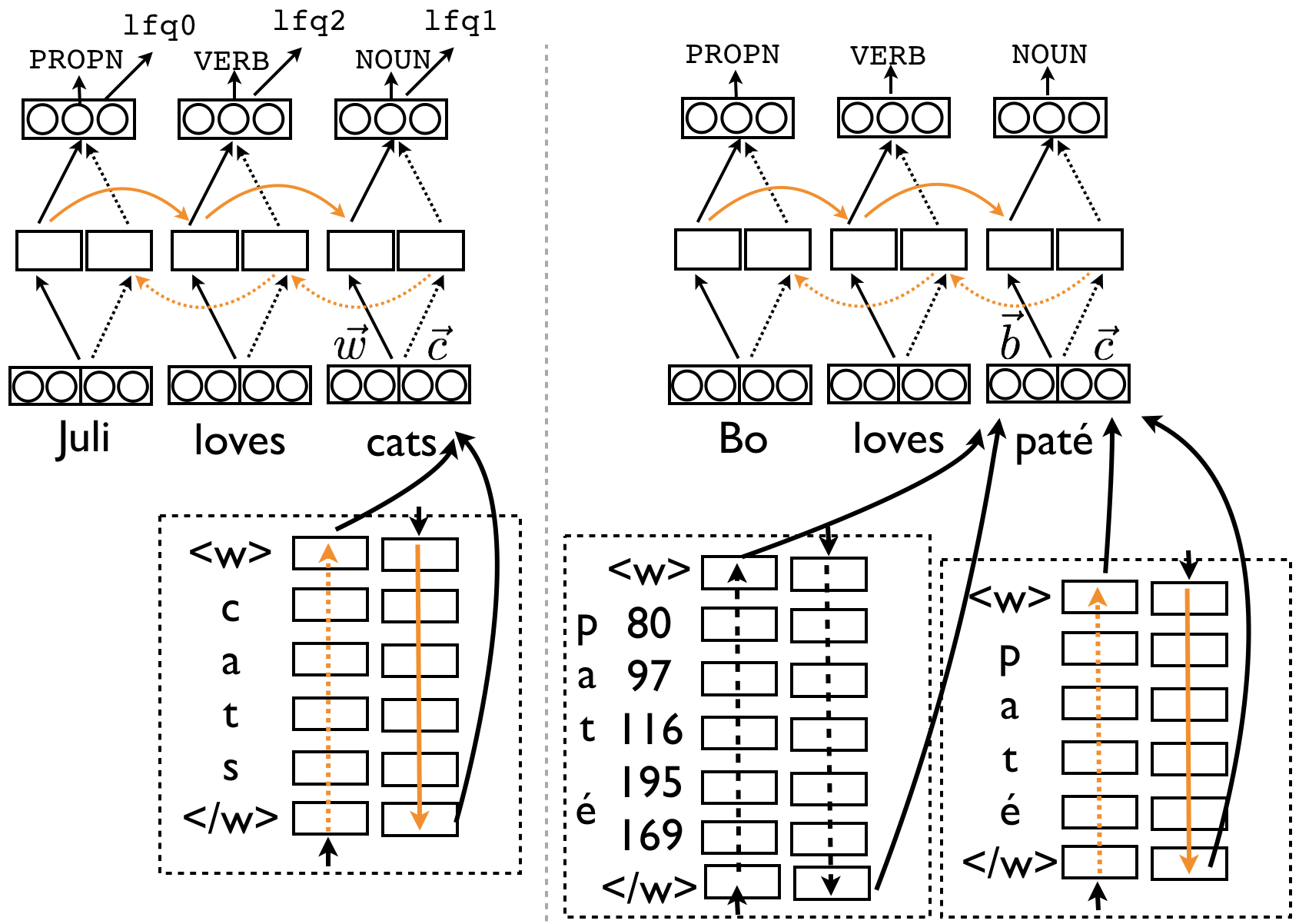}
\caption{Right: bi-LSTM, illustrated with $\vec{b}+\vec{c}$ (bytes and characters), for $\vec{w}+\vec{c}$  replace $\vec{b}$ with words $\vec{w}$. Left: \textsc{Freqbin}, our multi-task bi-LSTM that predicts at every time step the tag and the frequency class for the token.}
\label{fig:bilstm}
\end{figure}

Our basic bi-LSTM tagging model is a context bi-LSTM taking as input word embeddings $\vec{w}$. 
We incorporate subtoken information using an hierarchical bi-LSTM architecture~\cite{ling:ea:2015,ballesteros:ea:2015}. 
We compute subtoken-level (either characters $\vec{c}$ or unicode byte $\vec{b}$) embeddings of words 
using a sequence bi-LSTM at the lower level. This representation is then concatenated with the (learned) word 
embeddings vector $\vec{w}$ which forms the input to the context bi-LSTM at the next layer. 
This model, illustrated in Figure~\ref{fig:bilstm} (lower part in left figure), is inspired by~\newcite{ballesteros:ea:2015}. 
We also test models in which we only keep sub-token information, e.g., either both byte and character 
embeddings (Figure~\ref{fig:bilstm}, right) or a single (sub-)token representation alone. 

In our novel model, cf.\ Figure~\ref{fig:bilstm} left, we train the bi-LSTM tagger to predict both the tags of the sequence, as well as a label that represents the  log frequency of the token as estimated from the training data. Our combined cross-entropy loss is now: $L(\hat{y_t},y_t) + L(\hat{y_{a}},y_{a})$, where $t$ stands for a POS tag and $a$ is the log frequency label, i.e., $a=int(log(freq_{train}(w))$.
 Combining this log frequency objective with the tagging task can be seen as an instance of multi-task learning in which the labels are predicted jointly. 
The idea behind this model is to make the representation predictive for frequency, which encourages the model to not share representations between common and rare words, thus benefiting the handling of rare tokens.

\section{Experiments} 
All bi-LSTM models were implemented in CNN/pycnn,\footnote{\url{https://github.com/clab/cnn}} a flexible neural network library. For all models we use the same hyperparameters, which were set on English dev, i.e., SGD training with cross-entropy loss, no mini-batches, 20 epochs, default learning rate (0.1), 128 dimensions for word embeddings, 100 for character and byte embeddings, 100 hidden states and Gaussian noise with $\sigma$=0.2. As training is stochastic in nature, we use a fixed seed throughout. Embeddings are not initialized with pre-trained embeddings, except when reported otherwise. In that case we use off-the-shelf \texttt{polyglot} embeddings~\cite{polyglot:2013}.\footnote{\url{https://sites.google.com/site/rmyeid/projects/polyglot}} No further unlabeled data is considered in this paper. The code is released at: \url{https://github.com/bplank/bilstm-aux} 

\paragraph{Taggers} We want to compare POS taggers under varying conditions. We hence use three different types of taggers: our implementation of a bi-LSTM; \textsc{Tnt}~\cite{brants2000tnt}
---a second order HMM with suffix trie handling for OOVs. We use \textsc{Tnt} as it was among the best performing taggers evaluated in~\newcite{horsmann:ea:2015}.\footnote{They found TreeTagger was closely followed by HunPos, a re-implementation of TnT, and Stanford  and ClearNLP were lower ranked. In an initial investigation, we compared Tnt, HunPos and TreeTagger and found Tnt to be consistently better than Treetagger, Hunpos followed closely but crashed on some languages (e.g., Arabic).} We complement the NN-based and HMM-based tagger with a CRF tagger, using a freely available implementation~\cite{Plank:ea:14:coling} based on \texttt{crfsuite}. 

\subsection{Datasets}

For the multilingual experiments, we use the data from the Universal Dependencies project v1.2~\cite{universal1.2} (17 POS) with the canonical data splits. 
For languages with token segmentation ambiguity we use the provided gold segmentation. If there is more than one treebank per language, we use the treebank that has the canonical language name (e.g.,  \emph{Finnish} instead of \emph{Finnish-FTB}). 
We consider all languages that have at least 60k tokens and are distributed with word forms, resulting in 22 languages. 
We also report accuracies on WSJ (45 POS) using the standard splits~\cite{Collins:02,Manning:11}. The overview of languages is provided in Table~\ref{tbl:lang}.

\begin{table}
\resizebox{\columnwidth}{!}{
\begin{tabular}{lll|lll}
\toprule
  & \textsc{coarse} & \textsc{fine} &  & \textsc{coarse} & \textsc{fine}\\
\midrule
ar	& non-IE	& Semitic & he	& non-IE	& Semitic\\
bg	& Indoeuropean	& Slavic & hi	& Indoeuropean	& Indo-Iranian\\
cs	& Indoeuropean	& Slavic & hr	& Indoeuropean	& Slavic\\
da	& Indoeuropean	& Germanic  & id	& non-IE	&Austronesian \\
de	& Indoeuropean	& Germanic &  it	& Indoeuropean	& Romance\\
en	& Indoeuropean	& Germanic & nl	& Indoeuropean	& Germanic\\
es	& Indoeuropean	& Romance &  no	& Indoeuropean	& Germanic\\
eu	& Language isolate	&  & pl	& Indoeuropean	& Slavic\\
fa	& Indoeuropean	& Indo-Iranian & pt	& Indoeuropean	& Romance\\
fi	& non-IE	& Uralic & sl	& Indoeuropean	& Slavic\\
fr	& Indoeuropean	& Romance & sv	& Indoeuropean	& Germanic\\
\bottomrule
\end{tabular}
}
\caption{Grouping of languages.}
\label{tbl:lang}
\end{table}

\subsection{Results}

Our results are given in Table~\ref{tbl:results}. First of all, notice that \textsc{TnT} performs remarkably well across the 22 languages, closely followed by CRF. The bi-LSTM
tagger ($\vec{w}$) without lower-level bi-LSTM for subtokens falls short, outperforms the traditional taggers only on 3 languages. The bi-LSTM model clearly benefits from character 
representations. The model using characters alone ($\vec{c}$) works remarkably well, it improves over \textsc{TnT} on 9 languages (incl.\ Slavic and Nordic languages). The combined word+character representation
model is the best representation, outperforming the baseline on all except one language (Indonesian), providing strong results already without pre-trained embeddings. This model (${\vec{w}+\vec{c}}$) reaches the biggest improvement (more than +2\% accuracy) on Hebrew and Slovene. 
 Initializing the word embeddings (+\textsc{Polyglot}) with off-the-shelf language-specific embeddings further improves accuracy.  
 The only system we are aware of that evaluates on UD is~\newcite{gillick:ea:2016} (last column). However, note that these results are not strictly comparable as they use the earlier UD v1.1 version.

The overall best system is the multi-task bi-LSTM \textsc{freqbin} (it uses ${\vec{w}+\vec{c}}$ and \textsc{Polyglot} initialization for $\vec{w}$). 
While on macro average it is on par with bi-LSTM ${\vec{w}+\vec{c}}$, it obtains the best results on 12/22 languages, and it is successful in predicting POS for OOV tokens 
(cf.\ Table~\ref{tbl:results} \textsc{OOV Acc} columns), especially for languages like Arabic, Farsi, Hebrew, Finnish. 

We examined simple RNNs and confirm the finding of  \newcite{ling:ea:2015} that they performed worse than their LSTM counterparts. 
Finally, the bi-LSTM tagger is competitive on WSJ, cf.\ Table~\ref{tbl:WSJ}. 

\begin{figure}[h!]
\includegraphics[width=\columnwidth]{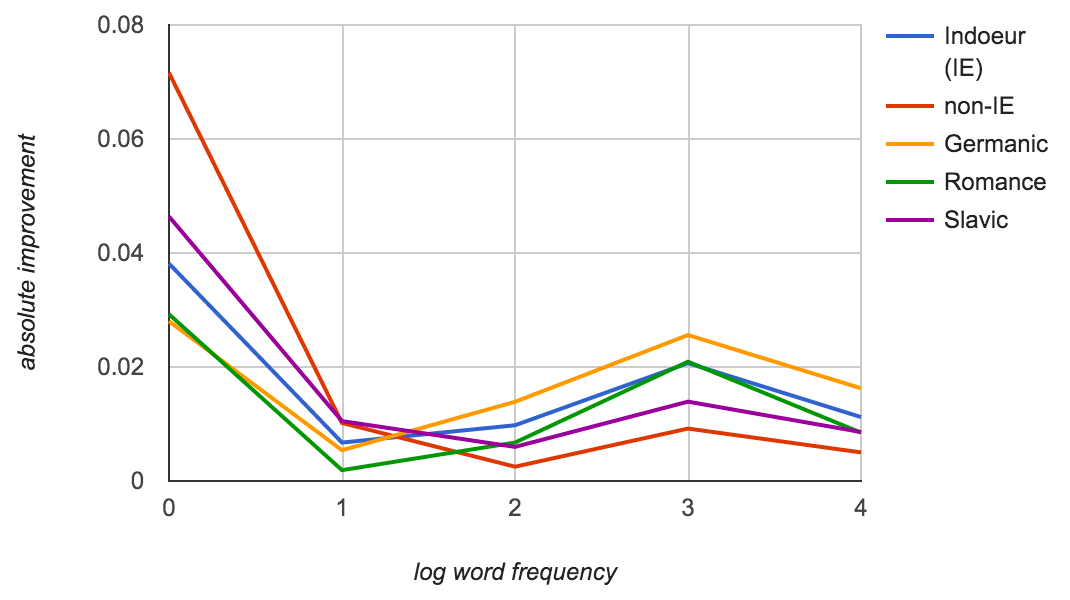}
\caption{Absolute improvements of bi-LSTM ($\vec{w}+\vec{c}$) over \textsc{Tnt} vs mean log frequency.}
\label{fig:logfreq}
\end{figure}

\paragraph{Rare words} In order to evaluate the effect of modeling sub-token information, we examine accuracy rates at different frequency rates. 
Figure~\ref{fig:logfreq} shows absolute improvements in accuracy of bi-LSTM $\vec{w}+\vec{c}$ over mean log frequency, for different language families. We see that especially for Slavic and non-Indoeuropean languages, having high morphologic complexity, most of the improvement is obtained in the Zipfian tail. Rare tokens benefit from the sub-token representations.

\begin{table*}[h!t]\centering
\resizebox{\textwidth}{!}{

\begin{tabular}{lcc|cccc|cc|cc|c}
\toprule
         & \multicolumn{2}{c}{\textsc{Baselines}} &  \multicolumn{4}{c}{\textsc{bi-LSTM} using:} &  \multicolumn{2}{c}{\textsc{${\vec{w}+\vec{c}}$ +Polyglot}} & \multicolumn{2}{c}{\textsc{OOV Acc}} & BTS\\
	& \textsc{Tnt} & \textsc{Crf} & $\vec{w}$ & $\vec{c}$ & $\vec{c}+\vec{b}$ & $\vec{w}+\vec{c} $  &  bi-LSTM & \textsc{Freqbin} &bi-LSTM & \textsc{Freqbin} & \\
\midrule
avg &94.61 & 94.27 & 96.00$\dagger$ & 94.29 & 94.01 & 92.37 & \textbf{96.50}  &\textbf{96.52}   & 83.48 &87.98 & 95.70\\
\midrule
Indoeur.      &  94.70  & 94.58  & 96.15$\dagger$ & 94.58 & 94.28 & 92.72 & \textbf{96.63} & \textbf{96.63} & 82.77 &  87.63& --\\
non-Indo.      & 94.57  & 93.62 & 95.67$\dagger$ & 93.51 &	93.16  & 91.97 & 96.21 & \textbf{96.28} & 87.44 &90.39  & --\\
Germanic        &  93.27& 93.21 & 95.09$\dagger$ & 92.89 & 92.59 & 91.18 & \textbf{95.55} & 95.49 & 81.22 & 85.45& --\\
Romance        &  95.37& 95.53 & 96.51$\dagger$ & 94.76 & 94.49 & 94.71 & \textbf{96.93} & \textbf{96.93} & 81.31 &86.07 & --\\
Slavic             &  95.64& 94.96  & 96.91$\dagger$  & 96.45 & 96.26 & 91.79 & 97.42 & \textbf{97.50} & 86.66 & 91.69& --\\
\midrule
ar	& 97.82 & 97.56 & \textbf{98.91} & 98.68 & 98.43 & 95.48 & 98.87 & \textbf{98.91} & 95.04 & 96.21 & --\\
bg	& 96.84 & 96.36 & 98.02 & 97.89 & 97.78 & 95.12 & \textbf{98.23} & 97.97 & 87.40 &  90.56&  97.84\\
cs	& 96.82  & 96.56& 97.80 & 96.38 & 96.08 & 93.77 & 98.02 & \textbf{98.24} & 89.02&  91.30& 98.50\\
da	& 94.29 & 93.83 & 96.19 & 95.12 & 94.88 & 91.96 & 96.16 & \textbf{96.35} & 77.09 & 86.35 &  95.52\\
de	& 92.64 & 91.38 & 92.64 & 90.02& 90.11 & 90.33 & \textbf{93.51} & 93.38 & 81.95& 86.77 & 92.87\\
en	& 92.66 & 93.35 & 94.46 & 91.62 & 91.57 & 92.10 & \textbf{95.17} & 95.16 & 71.23& 80.11 & 93.87\\
es	& 94.55 & 94.23 & 95.12 & 93.06 & 92.29 & 93.60 & 95.67 & \textbf{95.74} & 71.38 & 79.27 & 95.80\\
eu	& 93.35 & 91.63 &  94.70 & 92.48 & 92.72 & 88.00 & 95.38 & \textbf{95.51}& 79.87 & 84.30 & --\\
fa	& 95.98 & 95.65 &  97.19 & 95.82	 & 95.03 & 95.31 & \textbf{97.60} & 97.49& 80.00& 89.05  & 96.82\\
fi	& 93.59 & 90.32 &  94.85 & 90.25	 & 89.15 & 87.95 & 95.74 & \textbf{95.85}& 86.34 & 88.85 & 95.48\\
fr	& 94.51 & 95.14 &  95.80 & 94.39	 & 93.69 & 94.44 & \textbf{96.20} & 96.11& 78.09 &  83.54 & 95.75\\
he	& 93.71 & 93.63 &  95.79 & 93.74	 & 93.58 & 93.97 & 96.92 & \textbf{96.96}& 80.11 & 88.83 & --\\
hi	& 94.53 & 96.00 &  96.23 & 93.40 & 92.99 & 95.99 & 96.97 & \textbf{97.10}& 81.19& 85.27 & --\\
hr	& 94.06 & 93.16 &  94.76 &95.32 & 94.47 & 89.24 & 96.27 & \textbf{96.82}& 84.62& 92.71 & --\\
id	& 93.16 & 92.96 &  93.11 & 91.37 & 91.46 & 90.48 & 93.32 & \textbf{93.41}& 88.25 & 87.67 & 92.85\\
it	& 96.16 & 96.43 &  97.59 & 95.62 & 95.77 & 96.57 & 97.90 & \textbf{97.95}& 83.59 & 89.15 & 97.56\\
nl	& 88.54 & 90.03 &  93.32 & 89.11 & 87.74 & 84.96 & \textbf{93.82} & 93.30& 76.62 & 75.95 & --\\
no	& 96.31 & 96.21 &  97.57 & 95.87	 & 95.75 & 94.39 & \textbf{98.06} & 98.03& 92.05 & 93.72  & --\\
pl	& 95.57 & 93.96 &  96.41 & 95.80 & 96.19 & 89.73 & \textbf{97.63} & 97.62& 91.77 & 94.94 & --\\
pt	& 96.27 & 96.32 &  97.53 & 95.96	& 96.20 & 94.24 & \textbf{97.94} & 97.90& 92.16 & 92.33 & --\\
sl	& 94.92 & 94.77 &  \textbf{97.55} & 96.87	 & 96.77 & 91.09 & \textbf{96.97} & 96.84& 80.48& 88.94 & --\\
sv	& 95.19 & 94.45 &  96.36 & 95.57 & 95.50 &93.32 & 96.60 & \textbf{96.69}& 88.37 & 89.80 & 95.57\\
\bottomrule

\end{tabular}
}
\caption{Tagging accuracies on UD 1.2 test sets. $\vec{w}$: words, $\vec{c}$: characters, $\vec{b}$: bytes. 
Bold/$\dagger$: best accuracy/representation; \textsc{+Polyglot}: using pre-trained embeddings. \
\textsc{Freqbin}: our multi-task model. \textsc{OOV Acc}: accuracies on OOVs. 
BTS: best results in~\newcite{gillick:ea:2016} (not strictly comparable). }
\label{tbl:results}
\end{table*}

\begin{table}\centering
\begin{small}
\begin{tabular}{lc}
\toprule
{\sc Wsj} & Accuracy\\
\midrule
Convnet \cite{santos:zadrozny:2014} & 97.32\\
Convnet reimplementation \cite{ling:ea:2015} & 96.80\\
Bi-RNN \cite{ling:ea:2015} & 95.93\\
Bi-LSTM \cite{ling:ea:2015} & 97.36\\
\midrule
Our bi-LSTM $_{\vec{w}+\vec{c}}$ & 97.22\\
\bottomrule
\end{tabular}
\end{small}
\caption{Comparison POS accuracy on WSJ; bi-LSTM: 30 epochs, $\sigma$=$0.3$, no \textsc{Polyglot}.}
\label{tbl:WSJ}
\end{table}

\begin{figure}
\includegraphics[width=1.0\columnwidth]{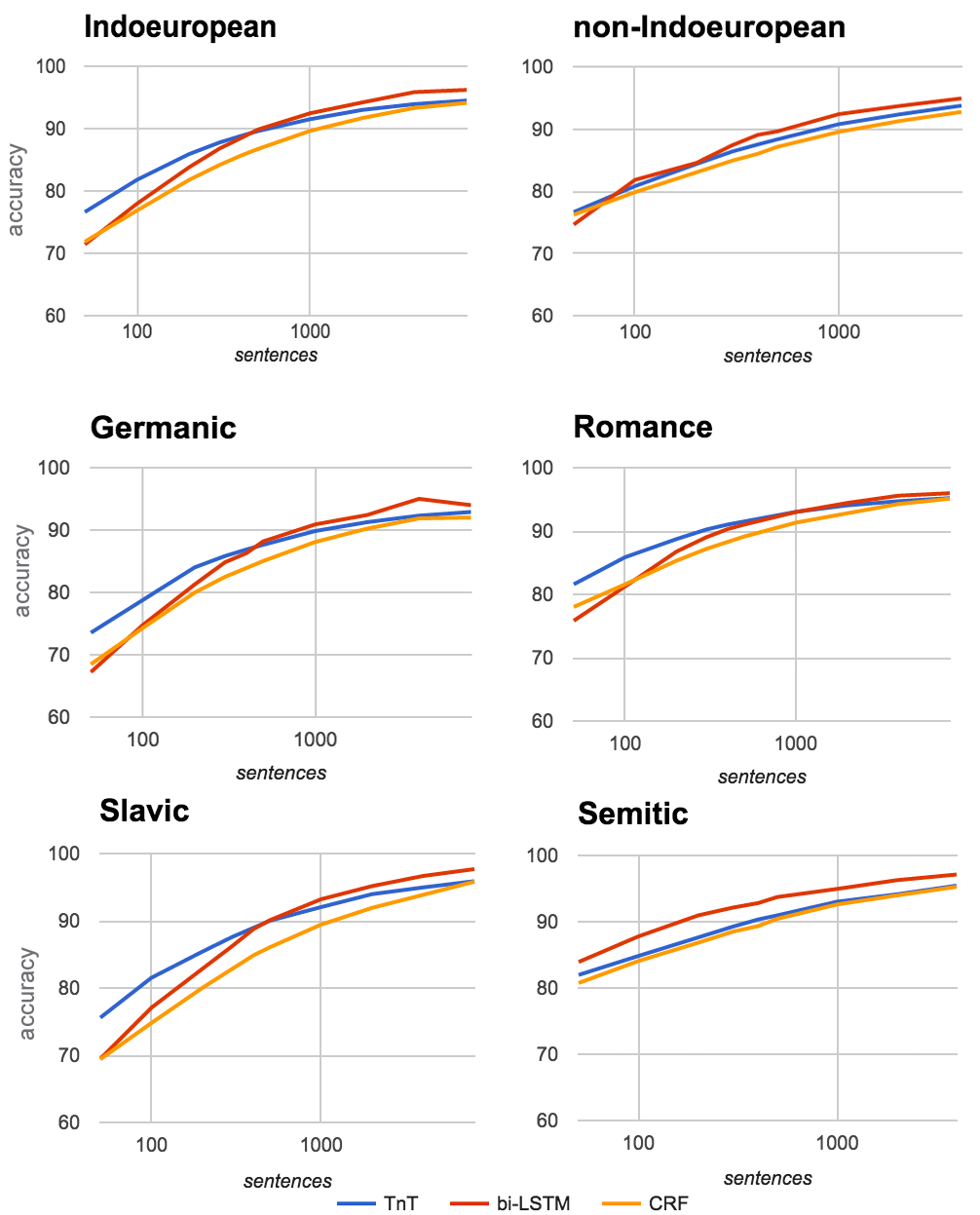}
\caption{Amount of training data (number of sentences) vs tagging accuracy. }
\label{fig:curve}
\end{figure}

\paragraph{Data set size} Prior work mostly used large data sets when applying neural network based 
approaches~\cite{zhang:ea:2015:char}. We evaluate how brittle such models are with respect to their more
traditional counterparts by training bi-LSTM ($\vec{w}+\vec{c}$ without Polyglot embeddings) for increasing amounts of training instances (number of sentences). The learning curves in Figure~\ref{fig:curve} 
show similar trends across language families.\footnote{We observe the same pattern with more, 40, iterations.} 
\textsc{TnT} is better with little data, bi-LSTM is better with more data, 
and bi-LSTM always wins over CRF. The bi-LSTM model performs already surprisingly well after only 500 training sentences. 
For non-Indoeuropean languages it is on par and above the other taggers with even less data (100 sentences). 
This shows that the bi-LSTMs often needs more data than the generative markovian model, but this is definitely less than what we expected.

\paragraph{Label Noise} We investigated the susceptibility of the models to noise, by artificially corrupting training labels. 
Our initial results show that at low noise rates, bi-LSTMs and \textsc{TnT} are affected similarly, their accuracies drop to a similar degree. Only at higher noise levels (more than 30\% corrupted labels), bi-LSTMs are less robust, showing higher drops in accuracy compared to \textsc{TnT}. This is the case for all investigated language families. 
\section{Related Work}

Character embeddings were first introduced by~\newcite{sutskever:ea:2011} for language modeling. Early applications include text classification~\cite{chrupala:2013,zhang:ea:2015:char}.  Recently, these representations were successfully applied to a range of structured prediction tasks. For POS tagging, 
\newcite{santos:zadrozny:2014} were the first to propose character-based models. 
They use a convolutional neural network (CNN; or convnet) and evaluated their model on English (PTB) and Portuguese, showing that the model achieves 
state-of-the-art performance close to taggers using carefully designed feature templates. \newcite{ling:ea:2015} extend this line and compare 
a novel bi-LSTM model, learning word representations through character embeddings. They evaluate their model on a language modeling and POS tagging setup,
and show that bi-LSTMs outperform the CNN approach of \newcite{santos:zadrozny:2014}. 
Similarly, \newcite{labeau:ea:2015} evaluate character embeddings for German. 
Bi-LSTMs for POS tagging are also reported in \newcite{wang:ea:2015:arxiv}, however, they only explore word embeddings, 
orthographic information and evaluate on WSJ only. 
A related study is \newcite{cheng:fang:ostendorf:2015} who propose a multi-task RNN for named entity recognition by jointly predicting the
next token and current token's name label. Our model is simpler, it uses a very coarse set of labels rather then integrating an entire language modeling
task which is computationally more expensive. An interesting recent study is~\newcite{gillick:ea:2016}, they build a single byte-to-span model for multiple
languages based on a sequence-to-sequence RNN~\cite{sutskever:ea:2014} achieving impressive results. We would like to extend this work in their direction.

\section{Conclusions}
We evaluated token and subtoken-level representations for neural network-based 
part-of-speech tagging across 22 languages and proposed a novel multi-task bi-LSTM
with auxiliary loss. The auxiliary loss is effective at improving the accuracy of rare words. 

Subtoken representations are necessary to obtain a state-of-the-art POS 
tagger, and character embeddings are particularly helpful for non-Indoeuropean and Slavic languages. 

Combining them with word embeddings in a hierarchical network provides the best representation.
The bi-LSTM tagger is as effective as the CRF and HMM taggers with already as little as 500 training sentences,
but is less robust to label noise (at higher noise rates). 

\section*{Acknowledgments}

We thank the anonymous reviewers for their feedback.  
AS is funded by the ERC Starting Grant
LOWLANDS No.\ 313695. YG is supported by The Israeli Science Foundation (grant number 1555/15) and a Google Research Award.

\bibliography{biblio}
\bibliographystyle{acl2016}

%
%
%
%


\end{document}